\documentclass[10pt,twocolumn,letterpaper]{article}

\usepackage[pagenumbers]{cvpr} 

\usepackage[dvipsnames]{xcolor}

\definecolor{cvprblue}{rgb}{0.21,0.49,0.74}
\usepackage[pagebackref,breaklinks,colorlinks,citecolor=cvprblue]{hyperref}

\title{Prim2Room: Layout-Controllable Room Mesh Generation from Primitives}

\author{Chengzeng Feng$^{1}$ \quad Jiacheng Wei$^{1}$ \quad Cheng Chen$^{1}$ \quad Yang Li$^{2}$ \quad Pan Ji$^{2}$ \\ \quad Fayao Liu$^3$ \quad Hongdong Li$^{4}$ \quad Guosheng Lin$^{1}$\vspace{0.3em} \\
{\normalsize $^1$Nanyang Technological University} \quad
{\normalsize $^2$Tencent XR Vision Labs, China} \quad \\
{\normalsize $^3$Institute for Infocomm Research A*STAR, Singapore} \quad
{\normalsize $^4$Australia National University}
}

\begin{document}
\maketitle

\begin{abstract}

We propose Prim2Room, a novel framework for controllable room mesh generation leveraging 2D layout conditions and 3D primitive retrieval to facilitate precise 3D layout specification. Diverging from existing methods that lack control and precision, our approach allows for detailed customization of room-scale environments. To overcome the limitations of previous methods, we introduce an adaptive viewpoint selection algorithm that allows the system to generate the furniture texture and geometry from more favorable views than predefined camera trajectories. Additionally, we employ non-rigid depth registration to ensure alignment between generated objects and their corresponding primitive while allowing for shape variations to maintain diversity. Our method not only enhances the accuracy and aesthetic appeal of generated 3D scenes but also provides a user-friendly platform for detailed room design.

\end{abstract}

\section{Introduction}
\label{sec:intro}

Recent years have seen significant advancements in 2D generative models, sparking increased interest in the potential of 3D generation technologies \cite{poole2022dreamfusion, gao2022get3d, lin2023magic3d, wang2023prolificdreamer}. However, these innovations have mostly focused on generating single objects or constructing relatively simple scenes. In contrast, this paper delves into the generation of room-scale 3D scenes. Although this area has received less attention in current research, it holds immense potential for various applications including VR/AR, interior design, and robotics.

\begin{figure}[h]
  \centering
  \includegraphics[width=\linewidth]{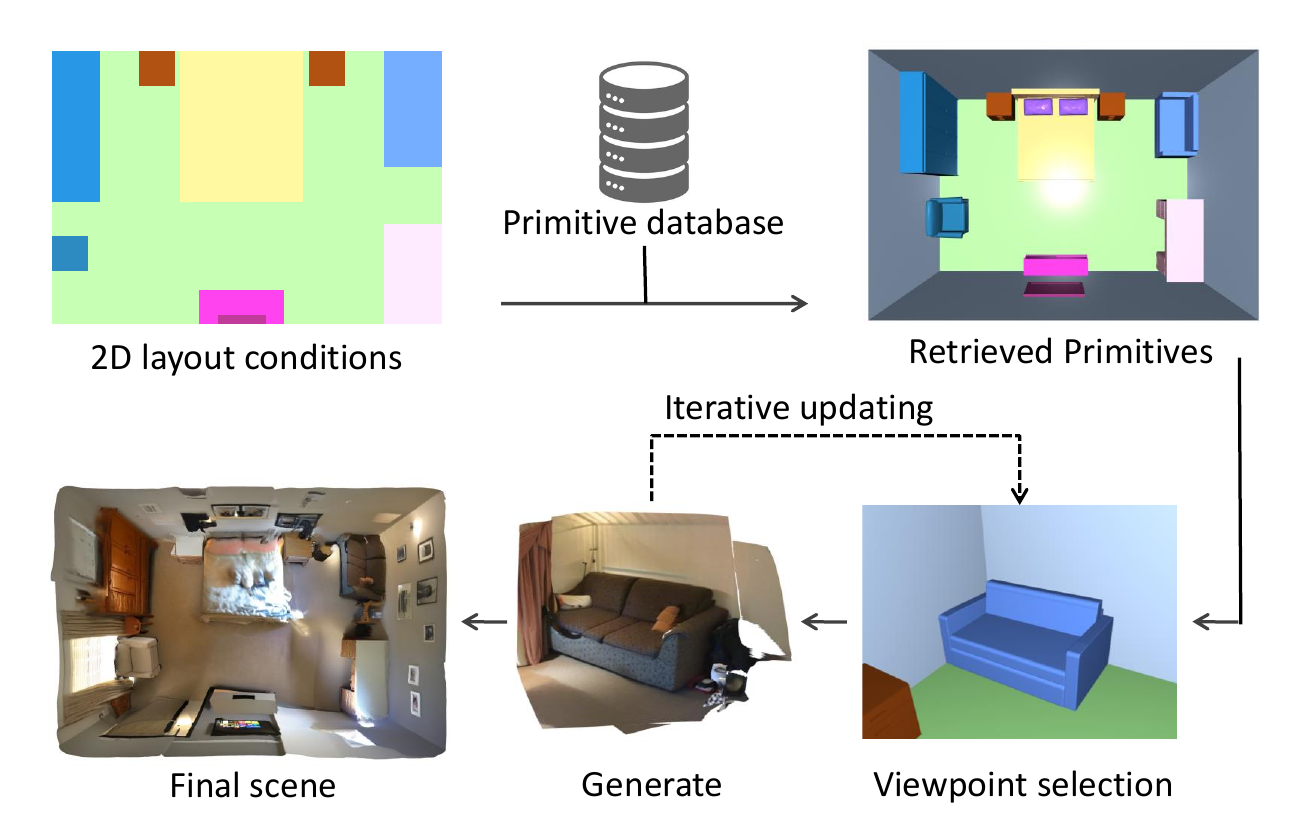}
  \caption{\textbf{Layout conditioned 3D room mesh generation.} We propose a room generation method that takes 2D bounding boxes as input conditions. We first retrieve a 3D primitive for each object, then create the room mesh through adaptive viewpoint selection and iterative mesh generation. Our method can generate compelling textures and geometry.
}
  \label{fig:overview}
\end{figure}

The recent work, Text2Room \cite{hollein2023text2room}, has significantly advanced the generation of textured 3D room meshes from text descriptions. This method uses an iterative inpainting process to fill in 2D regions, then enhances depth and 3D realism through fusion techniques. However, Text2Room primarily supports scene customization using only textual prompts, which may not provide enough precision for complex scenarios that require detailed layout control. Moreover, its dependency on pre-trained depth estimation models \cite{bae2022irondepth} can cause distortion in the final room geometry due to accumulated depth errors.

Building on these concepts, ControlRoom3D \cite{schult2023controlroom3d} aims to improve user control over the generation process by integrating semantic bounding boxes with text prompts, which allows more specific room layout customization. However, this approach presents unique challenges; creating 3D bounding boxes might be cumbersome for users and may not always express the desired level of detail, especially for intricate furniture styles or subcategory distinctions. This could potentially limit the granularity of user input and create a mismatch between the user's intentions and the model's output.

To enable a user-friendly and precise control mechanism over both the layout and the furniture shapes, we introduce a novel controllable 3D room mesh generation method that leverages the 2D bounding box specification and 3D primitive retrieval as a means to define 3D layout conditions. Users can specify each object's semantic category, size, and position with 2D bounding boxes. Along with a text prompt, our method retrieves a 3D primitive model from ShapeNet \cite{chang2015shapenet} for each bounding box, converting the 2D layout into 3D primitive representations.

Comparing with 3D bounding box conditions introduced in ControlRoom3D, our 2D layout specification and primitive retrieval scheme offers several distinct advantages. Firstly, 2D bounding boxes are easier for users to specify. The retrieved 3D primitives offer a more intuitive visualization approach, giving users the option to replace the retrieved primitive with other candidates if desired. Secondly, the geometric precision inherent in primitive models significantly enhances the system's ability to provide rich geometric cues, thereby improving the depth estimation process which is crucial for generating realistic 3D scenes.

While the 3D shapes retrieved from the primitive database can sometimes offer suitable geometry, in many cases, these primitive shapes are simply composed by several compact planes. This makes it challenging to meet the demand for diverse and natural scene synthesis. To create natural and photorealistic room meshes, we adopt an iterative scene generation approach. Different from existing methods \cite{hollein2023text2room, schult2023controlroom3d}, which usually use a predefined camera trajectory when generating any scene, we devise an Adaptive Viewpoint Selection (AVS) algorithm. For each object, this algorithm automatically identifies several favourable viewpoints for image generation and depth estimation models, which can prevent an object from being partially observed in multiple adjacent views.

Additionally, we observe that primitive shapes offer strong geometric guidance when using depth predictions to form mesh surfaces. Simple linear alignment between estimated scene depth and rendered primitive depth doesn't fully utilize geometric cues, as it only aligns depth prediction with conditional depth globally. To enable local shape deformation, some 3D registration works, like Neural Deformation Pyramids (NDP) \cite{li2022non}, formulate non-rigid registration as different levels of warping fields. However, directly using the NDP algorithm often causes texture distortion, as it fails to leverage pixel-to-pixel correspondence between shapes derived from two depth maps. To address this, we introduce Non-rigid Depth Registration (NDR), which constrains point movement along camera rays during shape deformation. Extensive experiments show that our method effectively warps depth predictions without texture distortion artifacts.

Our contributions can be concluded as: 
\begin{itemize}
    \item We propose Prim2Room, a controllable and photorealistic room mesh generation method which utilizes primitives for room layout definition, allowing users to precisely and intuitively dictate room boundaries and furniture arrangement.
    \item We introduce an adaptive viewpoint selection approach that automatically determines several views favorable for indoor scene image generation and depth prediction. This approach improves upon the predefined camera trajectories used in existing works.
    \item We devise a non-rigid depth registration algorithm, aligning generated scenes more closely with primitives compared to linear depth alignment. By adding a point movement constraint, we alleviate texture distortion artifacts existed in previous registration algorithms. 
\end{itemize}

\begin{figure*}[t]
  \centering
  \includegraphics[width=\textwidth]{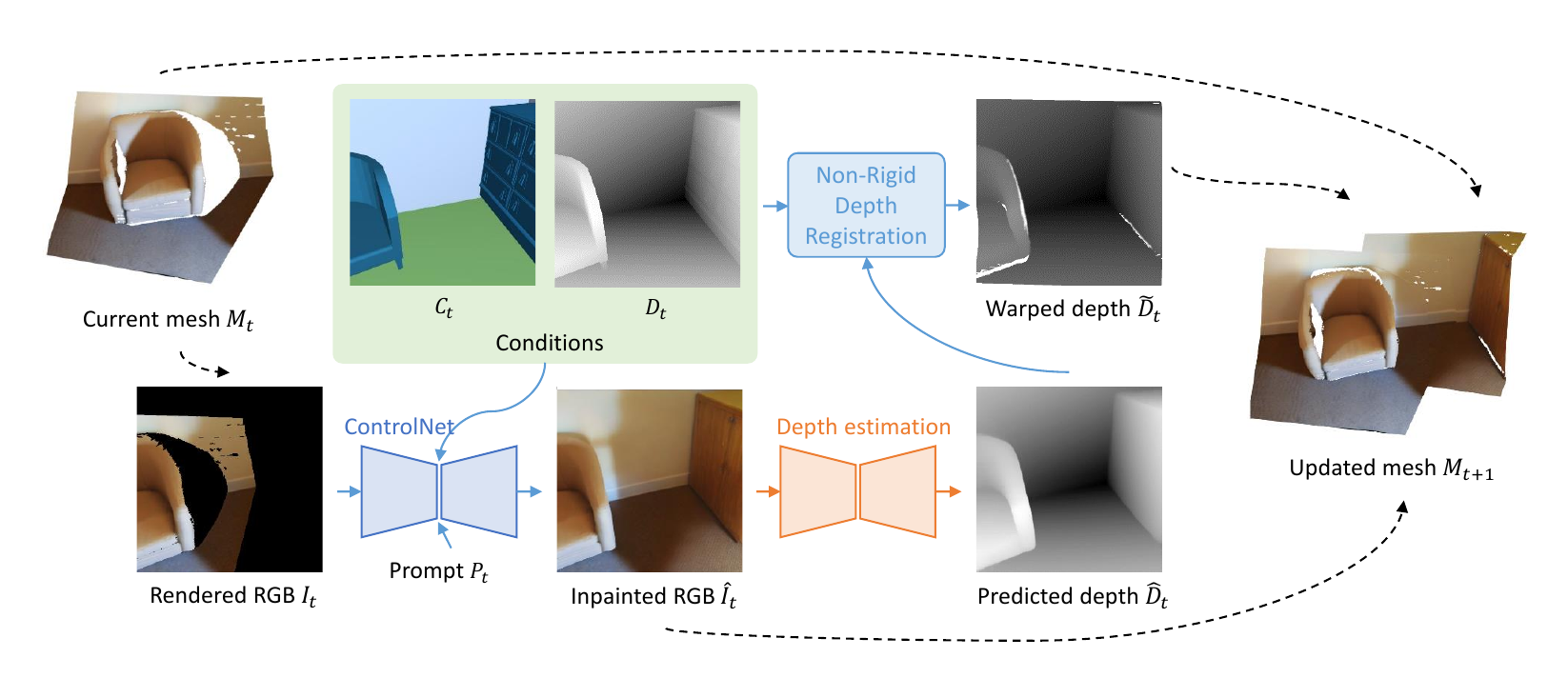}
  \caption{\textbf{Generation Pipeline.} Our method iteratively updates the scene mesh by a project-and-inpaint method. Before fusing the newly estimated frame into the existing mesh, we use a non-rigid depth registration method to fit the frame to both existing mesh and the conditioned primitives.  }
  \label{fig:pipeline}
\end{figure*}

\section{Related Work}

\subsection{3D indoor scene synthesis}

One group of 3D indoor scene generation works are retrieval-based methods \cite{paschalidou2021atiss, para2023cofs, tang2023diffuscene, feng2023layoutgpt}, these methods follow a retrieve then place pipeline which first retrieves CAD models from the 3D shape database and then regress the plausible size, orientation, and position for each CAD object to form the 3D scene. Since the CAD models are retrieved from the database, the scene diversity is limited by the predefined texture and geometry. 

Another type of works \cite{devries2021unconstrained, bautista2022gaudi, po2023compositional, son2023singraf, bahmani2023cc3d} generates 3D scenes using implicit representations like Neural Radiance Fields (NeRF) \cite{mildenhall2021nerf} or 3D Gaussian Splatting (3DGS) \cite{kerbl20233dgs}. While these methods can produce impressive novel-view images, they often struggle to extract high-quality 3D geometry. Additionally, many of these works rely on Generative Adversarial Networks (GANs) \cite{goodfellow2020generative} trained on datasets such as 3D-FRONT \cite{fu20213dfront} and Replica \cite{straub2019replica}, limiting their ability to generate diverse and photorealistic indoor scenes. Text2NeRF \cite{zhang2024text2nerf} and LucidDreamer \cite{chung2023luciddreamer} can achieve realistic effects, but they can hardly synthesize complete indoor scenes that allow for free-moving viewpoints.

Since explicit mesh representation has more practical application in areas such as VR/AR and games, there are also works \cite{hollein2023text2room, fang2023ctrl, schult2023controlroom3d, wu2024blockfusion} that directly generate room meshes. One representative work is Text2Room \cite{hollein2023text2room}, which creates local patches by iteratively synthesizing RGB-D frames and then fuses these frames to obtain the final mesh. To generate scenes based on rough room layouts, ControlRoom3D \cite{schult2023controlroom3d} extends Text2Room with semantic bounding boxes condition. However, their results tend to be more artistic than photorealistic. BlockFusion \cite{wu2024blockfusion} enables 2D layout control to synthesize room meshes, but they only generate the room geometry. Additionally, a series of works \cite{Tang2023mvdiffusion, chen2024scenetex, song2023roomdreamer, yang2024dreamspace} concentrate on generating realistic mesh textures for given room geometries.

\subsection{Non-rigid 3D registration}

Non-rigid 3D registration identifies the deformation that warps a source 3D shape to a target shape. Unlike rigid registration, which simply involves global rotation and translation, non-rigid registration is more complex. It requires estimating the unknown transformation of all points. 
To tackle the problem, existing works propose deformation graph \cite{sumner2007embedded, li2008global} to find the point-wise displacement, affine transformation with motion smoothness regularization \cite{liao2009modeling}, optimal transport with global bijective-matching constraint \cite{feydy2019interpolating}, coherent point drift \cite{hirose2020bayesian, myronenko2010point}. Transformer-based methods Lepard \cite{li2022lepard} further improves the result via learning a global point-to-point mapping. Besides the above-mentioned single-level methods, Neural Deformation Pyramid (NDP) \cite{li2022non} introduced a multi-level deformation model for better performance and faster inference. 

In this paper, we focus on solving the non-rigid registration between two point clouds derived from depth maps. Existing non-rigid methods allow excessive degrees of freedom for each point, potentially causing texture distortion artifacts. To address this, we constrain the movement of each point to the direction of the camera ray.

\section{Method}

Our method employs an iterative approach to generate the room mesh. We begin by automatically selecting camera viewpoints based on the layout conditions. For each view, we synthesize a realistic scene image using a ControlNet, which is conditioned by rendered primitive conditions. Next, we predict the scene depth using a depth estimator. We then align and register this predicted depth with the rendered primitive depth to enhance accuracy and consistency. Finally, we fuse the newly synthesized RGB-D contents to construct the updated mesh.

\subsection{Primitive Retrieval}

Given the 2D layout conditions, we first convert the 2D bounding boxes into 3D primitive representations through a simple retrieval process. We use ShapeNet \cite{chang2015shapenet} as our primitive database. For each bounding box, we select primitives that match its aspect ratio from the subset with the corresponding category. We also use CLIP \cite{radford2021learning} to find the target primitive if users provide a detailed description towards any single object.

\subsection{Viewpoint Selection}

Contrary to Text2Room \cite{hollein2023text2room}, which relies on a fixed camera trajectory for scene generation, our method employs adaptive selection of several favourable viewpoints for each object. We initiate this process by sampling candidate camera positions within the room space, then we introduce a scoring function $S(p_i)$ to evaluate the extent to which a candidate position $p_i$ is suitable as an observation position. 

\begin{equation}
  S(p_i)= S_{area} + w_i * S_{iou} + w_n * S_{norm}. 
  \label{eq:score_func}
\end{equation}

The first term \( S_{area} \) quantifies the visible surface area of the primitive shape when observed from position \( p_i \), which is oriented towards the primitive's geometric center. This term ensures that the selected viewpoint provides large visibility of the model.

The second term, \( S_{iou} \), computes the Intersection over Union (IoU) between the projected bounding box \( b_{proj} \) of the primitive and a predefined image range bounding box \( b_{range} = \left[(m, m), (w - m, h - m)\right] \), where \( h \), \( w \), and \( m \) denote the image height, width, and a predefined margin length, respectively. This measure encourages the selection of a camera position that maintains an appropriate distance from the object, ensuring the model is adequately framed within the image space.

Lastly, \( S_{norm} \) is introduced to evaluate the alignment between the primitive's surface normals and the inverse direction of camera rays, promoting orientations where the model's surfaces face the camera as directly as possible. Collectively, these components facilitate the selection of optimal camera positions for detailed and accurately oriented scene rendering.

We select several viewpoints for each primitive using an iterative approach. For the first view, $S_{area}$ calculates the visible surface area. For subsequent views, $S_{area}$ represents the newly observed surface area. We stop the process when there's no significant increase in newly observed surface area.

\subsection{Iterative Generation}

After viewpoint selection, we obtain a sequence of rendered condition maps $\{\left(C_t, D_t\right)\}_{t=1}^T$, where $C_t$ and $D_t$ represent the semantic and depth conditions, respectively. Additionally, we generate a view-dependent text prompt $P_t$ for each frame, determining which categories of objects are visible from the current viewpoint.

Drawing inspiration from \cite{hollein2023text2room, fridman2023scenescape}, we employ an iterative approach to synthesize the room mesh. At each generation step $t$, we render the current mesh $M_t$ to acquire the partial color map $I_t$. Subsequently, this partial color map is inpainted using a ControlNet \cite{zhang2023adding}, guided by the rendered conditions $C_t, D_t$ and the text prompt $P_t$:
\begin{equation}
  \hat{I}_t= \mathcal{F}_{control}(I_t, C_t, D_t, P_t). 
  \label{eq:controlnet}
\end{equation}
After that, we estimate the metric depth of the inpainted color map $\hat{I}_t$ using a pre-trained monocular depth estimation model \cite{yang2024depth}:
\begin{equation}
  \hat{D}_t= \mathcal{F}_{depth}(\hat{I}_t). 
  \label{eq:irondepth}
\end{equation}
We align and warp the estimated depth $\hat{D}_t$ to the depth condition $D_t$ through a scale-shift alignment estimation method and a non-rigid 3D registration algorithm, respectively:
\begin{equation}
  \widetilde{D}_t= \mathcal{F}_{warp}(\mathcal{F}_{align}(\hat{D}_t, D_t), D_t). 
  \label{eq:align_warp}
\end{equation}
Finally, we update the current mesh $ M_t $ with the synthesized RGB $ \hat{I}_t $ and warped depth $ \widetilde{D}_t $ into the current mesh:
\begin{equation}
  M_{t+1} = fuse(M_t, \hat{I}_t, \widetilde{D}_t). 
  \label{eq:fuse_mesh}
\end{equation}

We perform room generation in a two-stage process. In the first stage, we synthesize the foreground objects using adaptively selected viewpoints. After fusing multiple frames, we remove the background surfaces according to the segmentation masks inferred from the Segment Anything Model (SAM) \cite{kirillov2023segment}. In the subsequent stage, we generate the background walls and floors using a predefined camera trajectory, akin to the approach employed in Text2Room.

\subsection{ControlNet Training}

Based on Stable Diffusion \cite{rombach2022high}, ControlNet \cite{zhang2023adding} achieves high-quality image generation under flexible conditions. In this work, we train a ControlNet to synthesize indoor scene images with semantic condition $C_t$ and depth condition $D_t$.

Our training data are derived from the ScanNet \cite{dai2017scannet}, Scan2CAD \cite{avetisyan2019scan2cad}, and SceneCAD \cite{avetisyan2020scenecad} datasets. ScanNet is an indoor dataset comprising over a thousand scenes. Scan2CAD and SceneCAD provide additional annotations for most ScanNet scenes. From ScanNet, we obtain an RGB-D sequence captured by a depth sensor. Scan2CAD aligns a CAD model from ShapeNet \cite{chang2015shapenet} for each major instance, while SceneCAD annotates the planes for the floor, ceiling, and walls.

We render the CAD models and plane annotations using the camera parameters provided by ScanNet and create view-dependent text prompts. These renderings and prompts serve as the input conditions for training our specialized ControlNet, supervised by the corresponding RGB frames captured by the depth sensor.

\subsection{Alignment and Registration}

To lift the scene images synthesized by ControlNet into 3D space, we employ a monocular depth estimator \cite{yang2024depth} to derive pixel-wise metric depth $\hat{D}_t$. However, these estimated depth maps often lack accuracy due to scale ambiguity. 

We observe that conditional primitive shapes offer strong guidance on absolute scene depth. To leverage such guidance, we first conduct linear alignment between the estimated depth $\hat{D}_t$ and the rendered conditional depth $D_t$ by determining scale and shift parameters $ \gamma, \beta \in \mathbb{R} $ through least squares optimization:
\begin{equation}
  \underset{\gamma, \beta}{\text{min}} {\lVert \gamma \cdot \hat{D}_t + \beta - D_t \rVert} ^2.  
  \label{eq:scale_shift}
\end{equation}
However, linear alignment can only take advantage of conditional depth guidance at a coarse scale. Therefore, we perform non-rigid registration to ensure that the globally aligned depth adapts to the conditioned depth more closely at a local scale. 

\begin{figure*}[]
  \centering
  \includegraphics[width=0.95\textwidth]{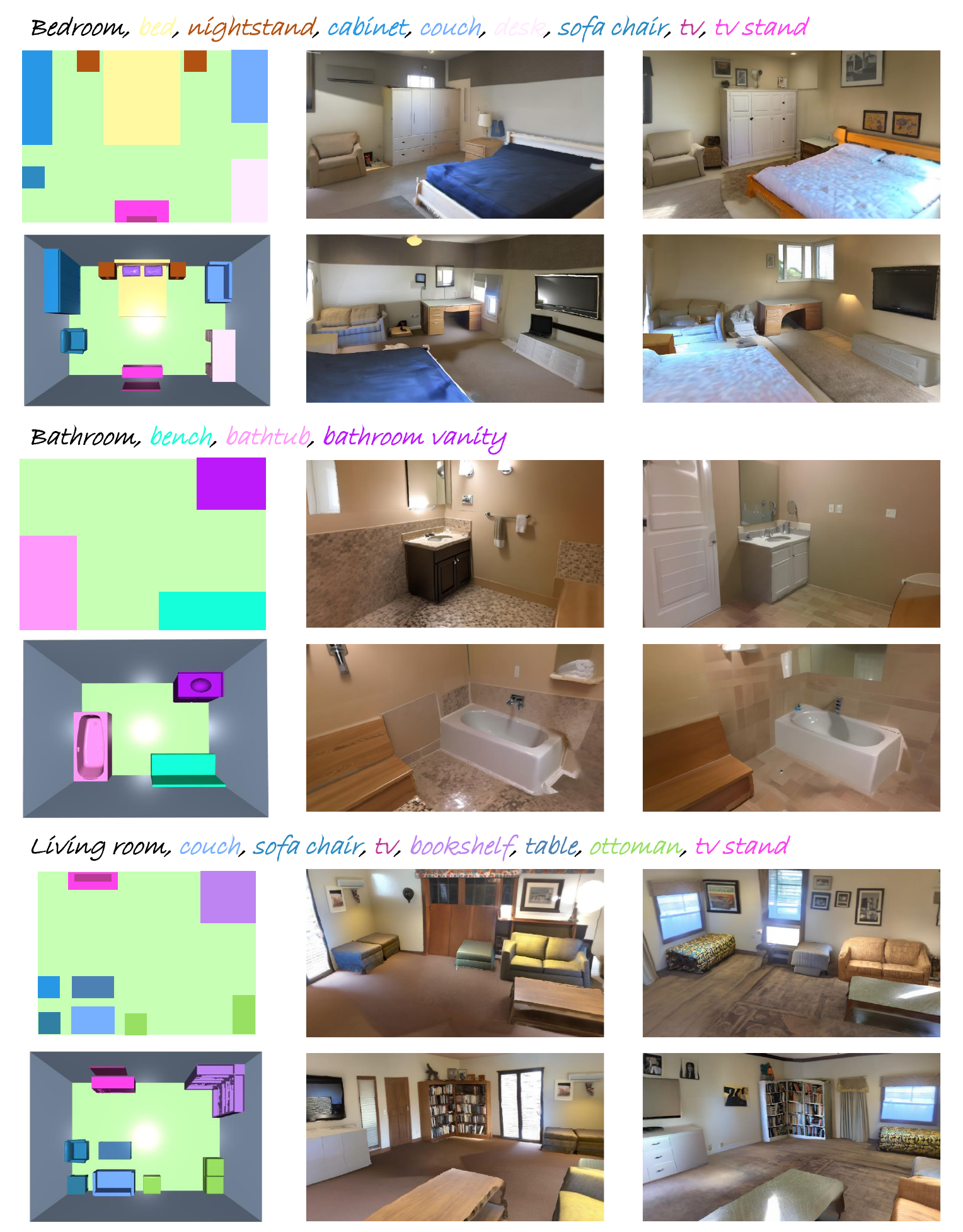}
  \caption{\textbf{Layout conditioned 3D room generation results of our method.} Given 2D layout conditions, we can generate high-quality room meshes consistent with the layout specification. }
  \label{fig:results}
\end{figure*}

\begin{figure*}[]
  \centering
  \includegraphics[width=\textwidth]{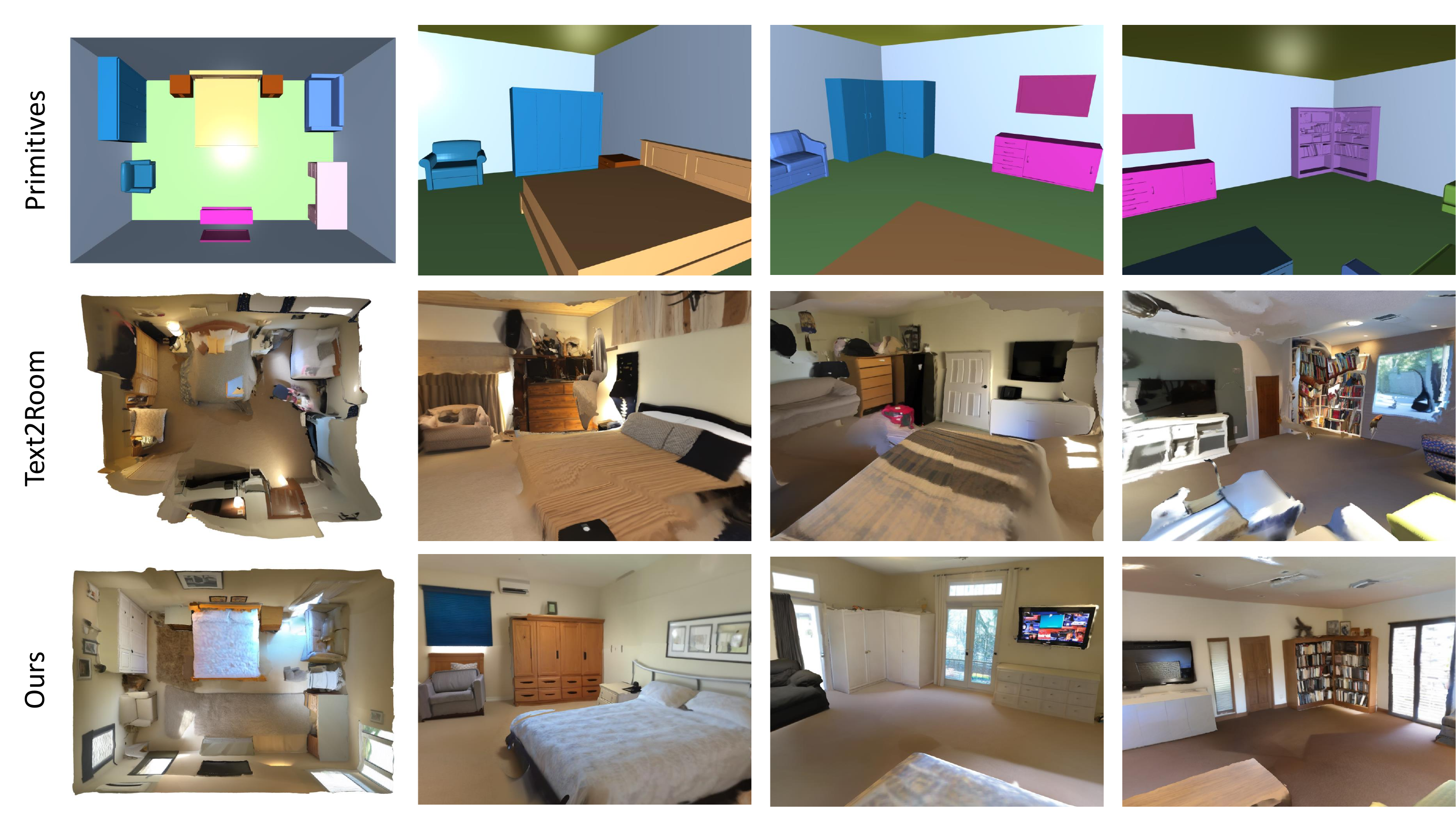}
  \caption{\textbf{Qualitative Comparison of Our Method and Baseline.} Comparing with Text2Room \cite{hollein2023text2room}, our method can generate room meshes more consistent with the retrieved primitives. Our results also demonstrate higher-quality room boundary and furniture shapes.
}
  \label{fig:compare_baseline}
\end{figure*}

Inspired by the concept of the Neural Deformation Pyramid (NDP) \cite{li2022non}, which uses several Multi-Layer Perceptrons (MLPs) to model the warping field hierarchically across different scales of shape deformation, we introduce a Non-rigid Depth Registration (NDR) technique specifically for shapes derived from depth maps. 

In the raw NDP algorithm, a rotation component $ R_i \in SO(3)$ and a translation vector $ t_i \in {\mathbb{R}}^3$ are estimated for each point $ x_i $ in the source point cloud to enable shape deformation: 
\begin{equation}
  \mathcal{W}_{ndp}(x_i; (R_i, t_i)) = R_i x_i + t_i.  
  \label{eq:ndp_warp}
\end{equation}
Each point has 6 degrees of freedom (DoF) in this warp function. Despite its impressive shape fitting ability, this method can often result in texture distortion due to incorrect correspondences when calculating the Chamfer Distance (CD) loss without constraining the point DoF during optimization.

To alleviate the texture distortion artifacts, we modify the warp function to impose a restriction where each point is only allowed to move along its corresponding camera ray. Therefore, our warp function is defined as:
\begin{equation}
  \mathcal{W}_{ndr}(x_i; {\delta}_i) = x_i + {\delta}_i \cdot r_i,  
  \label{eq:ndr_warp}
\end{equation}
where $r_i$ denotes the normalized camera ray vector, and ${\delta}_i \in \mathbb{R}$ is the variable to be estimated, representing the step size that point $p_i$ moves along the camera ray. 

To optimize the warp field at each level of the deformation pyramid, we employ the same loss function as NDP.

\section{Experiments}

\subsection{Implementation Details}

Color maps captured by ScanNet are typically characterized by noise and blurriness. To address this issue, we employ Real-ESRGAN \cite{wang2021realesrgan} to enhance the image quality prior to training the ControlNet. During the training phase of ControlNet, we implement data augmentation techniques such as random flips and crops to enrich the dataset diversity. Additionally, Depth Anything V2 \cite{yang2024depth} is utilized as our metric depth estimator.

\begin{figure*}[]
  \centering
  \includegraphics[width=\textwidth]{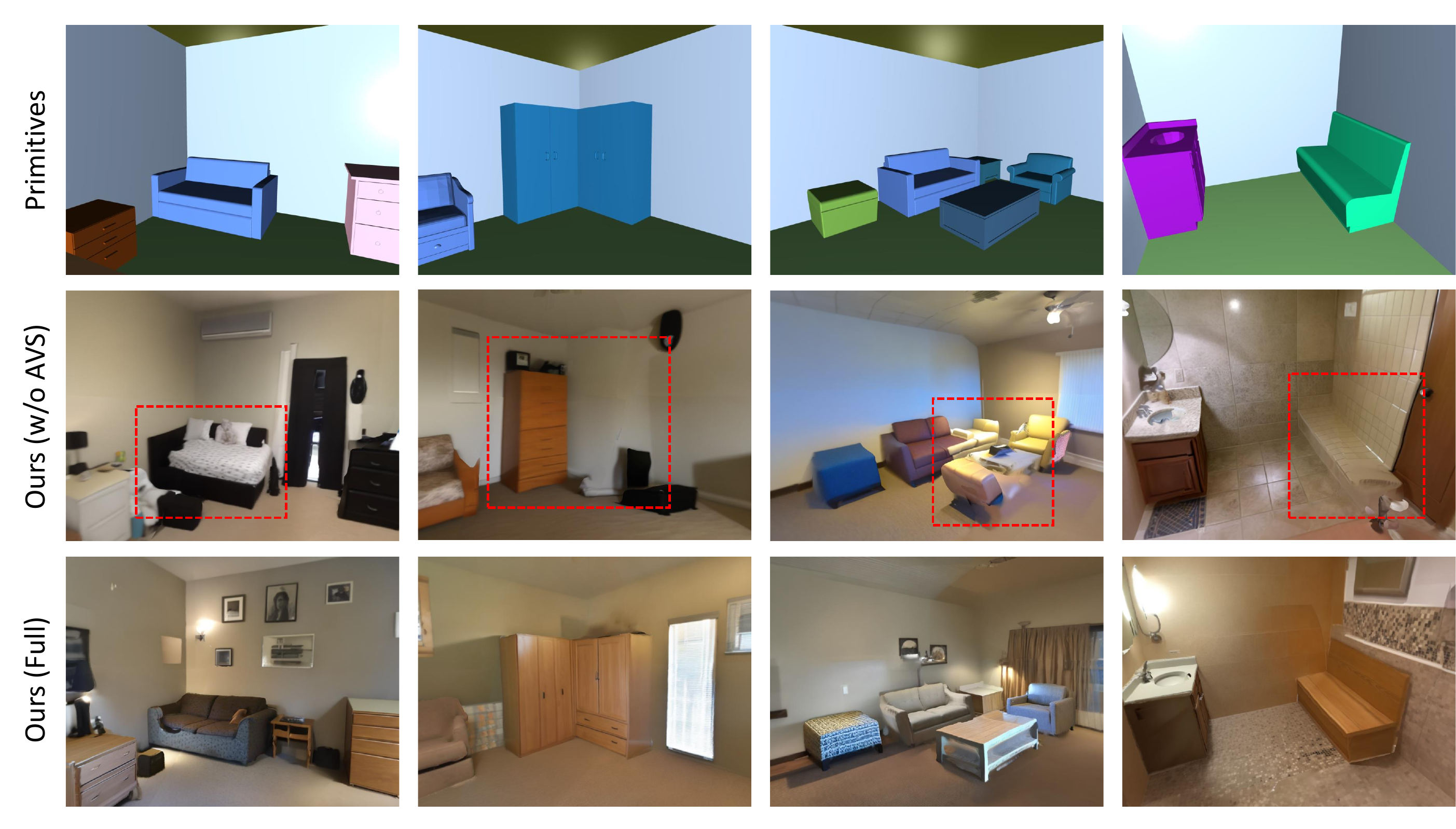}
  \caption{\textbf{Ablation Study on Viewpoint Selection.} Our proposed Adaptive Viewpoint Selection (AVS) algorithm helps generate contents more consistent with the retrieved primitives.
}
  \label{fig:ablation_avs}
\end{figure*}

\subsection{Quantitative Comparison}
\label{sec:quant}

\textbf{Evaluation Metric.} We evaluate the synthesized 3D scenes using both the 2D metric and user study. For the 2D metric, we calculate the CLIP Score (CS) \cite{hessel2021clipscore} for each scene using novel view RGB renderings. We also conduct a user study with 10 users across 35 scenes. In each scene, we display two 3D windows side-by-side showing the conditional layout specification (left) and the synthesized room mesh (right). Participants are asked to rate these scenes on a scale of 1–5 based on three factors: Layout Consistency (LC), geometric quality (GQ), and Perceptual Quality (PQ). 

\begin{table}[tb]
  \caption{\textbf{Quantitative Comparison.} We report the 2D metric CLIP Score (CS) and user study results, including Layout Consistency (LC), geometric quality (GQ) and Perceptual Quality (PQ).
  }
  \label{tab:quant_comp}
  \centering
  \begin{tabular}{@{}lcccc@{}}
    \toprule
    Methods & CS $\uparrow$ & LC $\uparrow$ & GQ $\uparrow$ & PQ $\uparrow$ \\
    \midrule
    Text2Room \cite{hollein2023text2room}  & 27.86 & 3.04 & 2.50 & 2.81\\
    Ours (w/o AVS) & 28.21 & 3.48 & 3.80 & 3.93\\
    Ours (w/o NDR) & 27.69 & 3.75 & 3.66 & 3.66\\
    Ours (raw NDP) & 27.94 & 4.06 & 3.46 & 3.94\\
    Ours (full) & {\bf 28.22}  & {\bf 4.60} & {\bf 4.06} & {\bf 4.46}\\
  \bottomrule
  \end{tabular}
\end{table}

\textbf{Results.} Tab.~\ref{tab:quant_comp} presents the results of the quantitative comparison. A key discovery is that our proposed Adaptive Viewpoint Selection (AVS) module and Non-rigid Depth Registration (NDR) module make significant contributions to layout consistency and geometric quality respectively. The layout consistency decreases by 1.12 when AVS is removed, and the geometric quality drops by 0.4 when NDR is removed.

The results also indicate that replacing the NDR module with the raw Neural Deformation Pyramid (NDP) algorithm \cite{li2022non} decreases both the geometric and overall quality, demonstrating that our NDR method can mitigate the distortion artifacts caused by NDP.

\subsection{Qualitative Results}

Figure \ref{fig:results} illustrates various types of room generation results achieved by our method. We are able to obtain high-quality and diverse textured meshes that closely align with the conditioned spatial layouts. 

\begin{figure*}[]
  \centering
  \includegraphics[width=\textwidth]{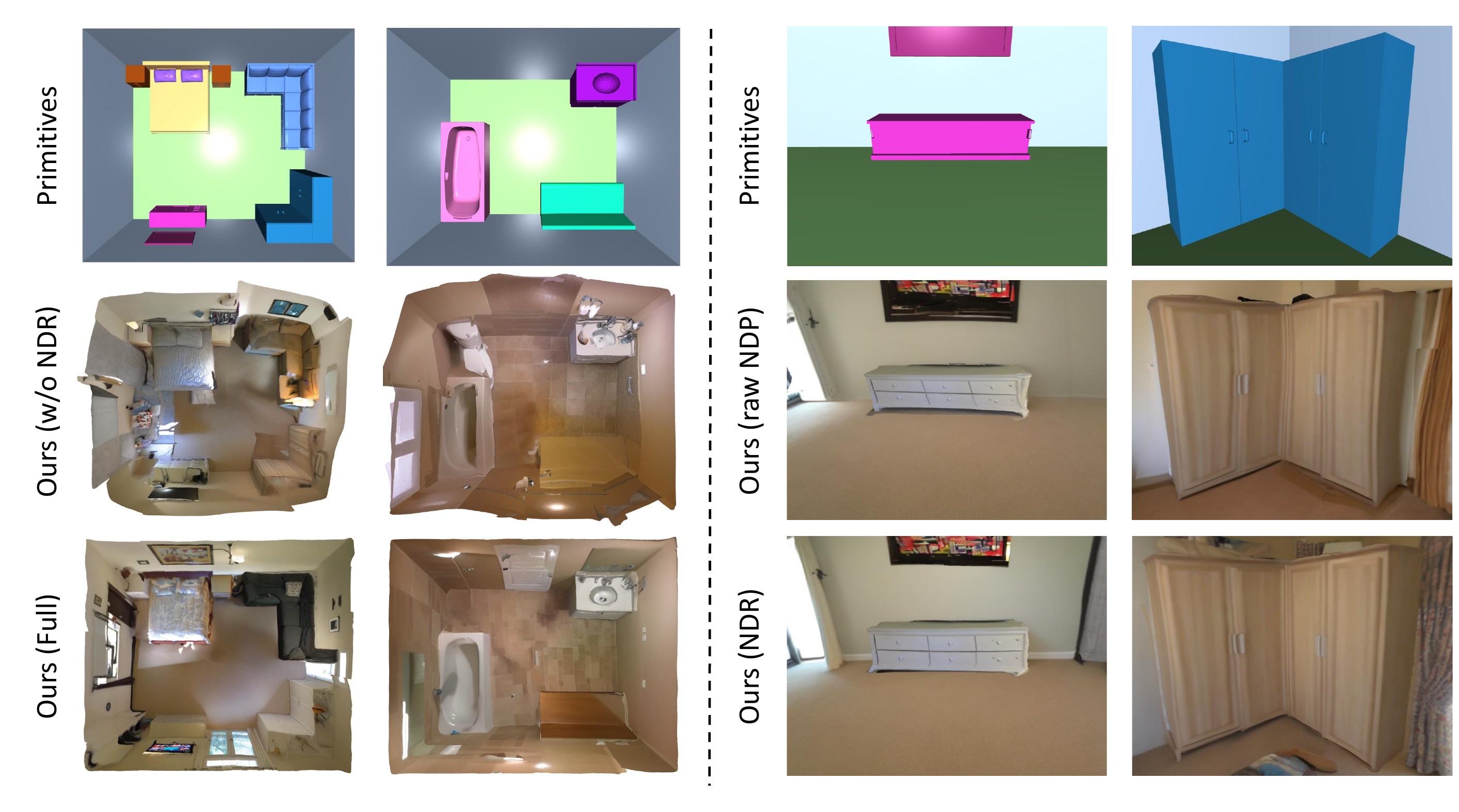}
  \caption{\textbf{Ablation Study on Non-rigid Depth Registration.} Compared to linear alignment, our Non-rigid Depth Registration (NDR) method creates flatter walls. NDR can also reduce texture distortion artifacts better than existing 3D registration algorithms such as Neural Deformation Pyramids (NDP) \cite{li2022non}.
}
  \label{fig:ablation_ndr}
\end{figure*}

We compare our method against the most closely related work, Text2Room \cite{hollein2023text2room}, which originally employs a text-conditioned inpainting model followed by monocular depth inpainting to iteratively generate the whole room. In our experiments, to make fair comparison, we adapt Text2Room by replacing its Stable Diffusion module with our primitive-conditioned ControlNet, which we trained independently.

Experimental results shown in Figure \ref{fig:compare_baseline} indicate that Text2Room tends to create unusual room boundary shapes due to the accumulated error in depth estimation. In contrast, our method is capable of synthesizing rooms with flat walls and floors, supported by our depth alignment and registration process. Additionally, the quality of furniture representation is notably enhanced in our results compared to the baseline.

\subsection{Ablation Study}

In this subsection, we evaluate the contribution of two main parts, adaptive viewpoint selection and non-rigid depth registration. 

Figure \ref{fig:ablation_avs} demonstrates that our proposed Adaptive Viewpoint Selection (AVS) is effective in generating scenes consistent with conditional primitives. When we drop AVS, primitives are often partially observed in adjacent predefined viewpoints, making it challenging for ControlNet to generate scene images that align with these primitives. In contrast, our AVS method observes primitives from appropriate distances and angles, leading to high-quality scene synthesis.

In Figure \ref{fig:ablation_ndr}, we illustrate the effects of our refined Non-rigid Depth Registration (NDR) module. Without NDR, depth estimations are only linearly aligned with conditions, often resulting in unrealistic room geometry, particularly for walls and floors. If we replace NDR with the raw Neural Deformation Pyramids (NDP) algorithm, distinct texture distortions can often be observed. 

\section{Conclusion}

In this paper, we present a controllable textured 3D room mesh generation method that empowers users to define the scene layout through 2D bounding boxes. We introduce an adaptive viewpoint selection method alongside a non-rigid depth registration algorithm, which, when combined, enable the systematic generation of indoor scenes characterized by high-quality textures and geometric precision. 

\textbf{Limitations.} Due to limitations in depth estimator accuracy, our system sometimes struggles to generate complex geometric structures. This is particularly evident with thin objects like chair legs and intricate shapes such as potted plants.

{
    \small
    \bibliographystyle{ieeenat_fullname}
    \bibliography{main}
}

\end{document}